\documentclass[letterpaper,10pt,conference]{ieeeconf}
\usepackage{url}
\makeatletter
\g@addto@macro\UrlBreaks{\do\a\do\b\do\c\do\d\do\e\do\f\do\g\do\h\do\i\do\j\do\k\do\l\do\m\do\n\do\o\do\p\do\q\do\r\do\s\do\t\do\u\do\v\do\w\do\x\do\y\do\z\do\A\do\B\do\C\do\D\do\E\do\F\do\G\do\H\do\I\do\J\do\K\do\L\do\M\do\N\do\O\do\P\do\Q\do\R\do\S\do\T\do\U\do\V\do\W\do\X\do\Y\do\Z\do\0\do\1\do\2\do\3\do\4\do\5\do\6\do\7\do\8\do\9}
\makeatother
\usepackage{amsmath}
\usepackage{graphicx}
\usepackage{booktabs}
\usepackage{multirow}
\usepackage[protrusion=false]{microtype}
\usepackage[letterpaper,margin=1.05in,includefoot,footskip=20pt]{geometry}
\renewcommand{\footnotesize}{\fontsize{10}{12}\selectfont}
\renewcommand{\small}{\fontsize{10}{12}\selectfont}
\renewenvironment{abstract}
  {\normalfont\normalsize\noindent\textbf{Abstract.}\ \ignorespaces}
  {\par\medskip}

\IEEEoverridecommandlockouts

\title{\fontsize{14}{17}\selectfont\bfseries
SimPT: An Automated Simulation Framework for\\
Intervention-Based VLA Post-Training
}

\author{\textbf{Caicheng Wang, Zipeng Ye$^{*}$, Zhexuan Zhou}\\
\textbf{Zilong Huang, Xi Zhang, Yuchen Xie}\\[4pt]
{\normalfont\fontsize{10}{12}\selectfont Meituan}%
\thanks{$^{*}$Corresponding author.}}

\begin{document}
\bstctlcite{IEEEcontrol}

\maketitle
\thispagestyle{plain}
\pagestyle{plain}
\raggedbottom
\clubpenalty=10000
\widowpenalty=10000

\begin{abstract}
Human-in-the-loop (HIL) post-training improves vision--language--action
(VLA) policies through expert corrections, but costly human supervision
and repeated real-robot evaluation limit algorithm exploration and
development. We introduce SimPT (Simulation-based Post-Training), a
simulation framework that automates intervention-based VLA post-training
by integrating policy rollouts, corrective data collection, policy
training, and standalone evaluation. To automatically determine
intervention timing and recovery strategies, we develop an Agent-Compiled
Progress Graph that tracks task progress using privileged simulator state
and retains stable snapshots during policy execution. After a rollout
fails, a CuRobo-based expert attempts recovery from these snapshots,
starting at more advanced task stages and falling back to earlier ones as
needed to generate corrective trajectories. We implement and compare four
post-training methods, RECAP, HG-DAgger, DPO+SFT, and FlowPRO, across seven
RoboTwin manipulation tasks. Relative to a fixed-interval snapshot
baseline, the Progress Graph reduces recovery attempts by 75.0\% and
increases recovery success from 93.4\% to 98.9\%. Using SimPT, we develop
RECAP-CF-LPF, a refinement of RECAP that combines cross-fitting and
low-pass filtering of value predictions. This refinement increases the
overall success rate from 76.50\% to 79.36\% in held-out simulation trials.
In real-robot HIL post-training, it increases the success rate from 72\%
to 96\% across 25 scenarios of an Ethernet cable-insertion task. These
results demonstrate the utility of SimPT for exploring and developing
VLA post-training algorithms.
\end{abstract}

\section{Introduction}
\label{sec:introduction}

Generalist vision--language--action (VLA) policies map visual observations
and language instructions directly to robot actions
\cite{black2025pi0,physicalintelligence2025pi05}. When these policies
encounter states outside the demonstration distribution, or
out-of-distribution (OOD) states, they may produce inappropriate actions
\cite{laskey2017dart}. Such errors can drive the robot further from the
demonstration distribution and compound over time~\cite{ross2011dagger}.
Human-in-the-loop (HIL) robot learning addresses this problem through expert
takeover, using corrective demonstrations to update the policy
\cite{ravichandar2020lfd,kelly2019hgdagger,luo2025hilserl}. However, developing
HIL algorithms on physical robots requires substantial human effort to
collect corrections and repeated hardware use for evaluation. Environmental
variation further complicates reproducible comparisons
\cite{li2025simpler}. These costs and sources of variability limit systematic
exploration and development of HIL post-training algorithms.

Simulation offers controllable scenes, repeatable trials, and scalable data
collection for developing and evaluating HIL algorithms before hardware
experiments
\cite{mandlekar2023mimicgen,jiang2025dexmimicgen}.
Platforms such as RoboTwin \cite{mu2025robotwin} and SIMPLER
\cite{li2025simpler} support demonstration generation, policy evaluation, and
sim-to-real research, while automated HIL post-training remains comparatively
underexplored. IntervenGen extends simulation-based data generation to
corrective behavior by expanding a small set of human interventions into
synthetic recovery trajectories, but relies on human experts to determine
when and how to intervene in the initial recovery demonstrations
\cite{hoque2024intervengen}.
Automating these decisions in simulation requires selecting suitable recovery
states along a failed policy rollout and determining how an expert should
complete the task from those states. Both depend on task progress.
Vision-language models (VLMs) can monitor execution, but repeated visual
queries incur inference overhead \cite{guo2024doremi}. CuRobo-based scripted
experts can execute predefined task sequences from initial states
\cite{sundaralingam2023curobo}, but the motion planner itself does not infer
task progress or select corrective actions for states reached by the policy.

\begin{figure*}[!t]
    \centering
    \setlength{\abovecaptionskip}{1pt}
    \includegraphics[width=\textwidth]{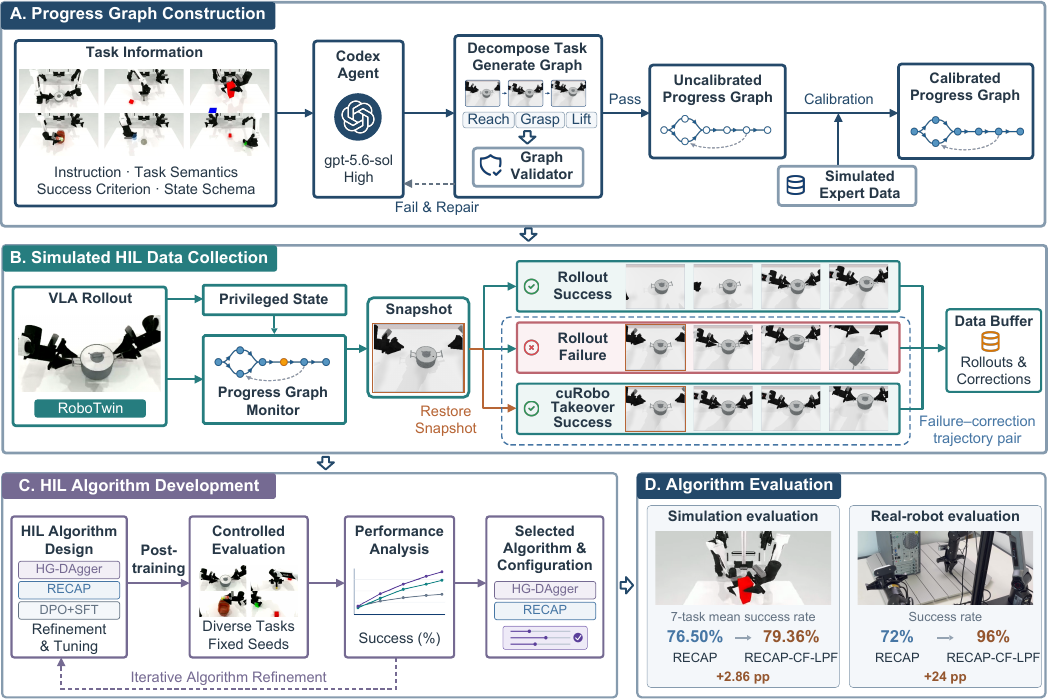}
    \caption{SimPT integrates corrective data collection, VLA post-training,
    and standalone policy evaluation in an automated simulation framework.
    (A) An agent constructs and validates a Progress Graph, whose
    data-dependent thresholds are calibrated using simulated expert trajectories.
    (B) After terminal rollout failure, stable snapshots enable expert recovery
    and corrective data collection. (C) Post-training and controlled evaluation
    support algorithm comparison and refinement; curves are schematic.
    (D) RECAP and RECAP-CF-LPF are compared using seven-task mean success
    rates on held-out simulation trials and Ethernet cable-insertion success rates
    on a real robot. Gains are reported in percentage points (pp).}
    \label{fig:framework}
\end{figure*}

Motivated by these challenges, we introduce \emph{SimPT}, a simulation
framework that integrates policy rollouts, automated corrective data
collection, policy training, and
standalone evaluation to support the exploration and development of VLA
post-training algorithms (Fig.~\ref{fig:framework}). To guide recovery, an agent
compiles task information into an \emph{Agent-Compiled Progress Graph}.
During policy execution, the graph tracks task progress through fixed checks
on privileged simulator state and retains stable snapshots without further
agent queries. After rollout failure, SimPT attempts expert recovery from
snapshots at more advanced task stages and falls back to earlier ones as
needed. A CuRobo-based expert resumes
execution from each restored state, and successful recovery trajectories
are retained for policy post-training.

Using SimPT, we implement and compare HG-DAgger \cite{kelly2019hgdagger},
RECAP \cite{physicalintelligence2025recap}, DPO+SFT \cite{rafailov2023dpo}, and
FlowPRO \cite{wu2026flowpro} on seven RoboTwin manipulation tasks. In
independent data collections, the Progress Graph requires 75.0\% fewer
recovery attempts than a fixed-interval snapshot baseline. We further use
SimPT to develop RECAP-CF-LPF, a refinement of RECAP that combines
cross-fitting (CF) and Butterworth low-pass filtering (LPF) of value
predictions. This combined refinement increases the overall
success rate from 76.50\% to 79.36\% in held-out simulation trials. Applied
to real-robot HIL post-training, it increases success from 72\% to 96\%
across 25 scenarios of an Ethernet cable-insertion task.
Our contributions are threefold:
\begin{itemize}
    \interlinepenalty=10000 
    \item We introduce SimPT, a simulation framework that automates the
    intervention-based VLA post-training workflow to support systematic
    algorithm exploration and development.

    \item We develop the Agent-Compiled Progress Graph to track task progress
    and retain stable snapshots during policy execution. After rollout failure,
    expert recovery starts from more advanced task stages and falls back to
    earlier ones as needed.

    \item We compare four post-training methods on seven RoboTwin tasks
    and use SimPT to develop RECAP-CF-LPF. By combining cross-fitting and
    low-pass filtering, this refinement improves policy success in both
    simulation and real-robot HIL post-training.
\end{itemize}

\section{Related Work}
\label{sec:related_work}

\subsection{Intervention-Based Robot Learning}

Interactive imitation learning mitigates distribution shift by collecting
expert supervision during policy execution \cite{zhang2017safedagger}.
DAgger aggregates expert labels on states encountered by the learner and
retrains the policy on the expanded dataset \cite{ross2011dagger}.
HIL variants differ in how they allocate expert control: HG-DAgger lets a
human expert decide when to take over and return control
\cite{kelly2019hgdagger}, whereas ThriftyDAgger requests assistance based on
novelty and estimated risk within an intervention budget
\cite{hoque2022thriftydagger}. Their behavior-cloning policy updates use
expert-labeled data \cite{celemin2022iilreview}, without directly using the
success or failure of autonomous rollouts as learning signals.

To learn from autonomous experience as well as expert corrections, recent
VLA post-training methods incorporate reinforcement learning (RL)
\cite{deng2025rlvlasurvey}. RECAP learns a value function to estimate future
returns, accounting for remaining task steps and failure penalties, and uses
the resulting advantages to condition policy training
\cite{physicalintelligence2025recap}. Noisy frame-level value estimates can
nevertheless yield misleading advantages in OOD states. STEAM addresses this
challenge by learning temporal-offset predictors from frame pairs in expert
demonstrations and conservatively combining their advantage estimates
\cite{liu2026steam}. Its ensemble reduces advantage overestimation at the
cost of training and evaluating multiple predictors.

Preference optimization offers an alternative without a separately trained
value function or critic. FlowPRO adapts direct preference optimization
(DPO) \cite{rafailov2023dpo} to flow-matching VLAs and uses a proximal
regularizer to constrain implicit reward magnitudes and mitigate reward
hacking. Its preference pairs are constructed from failed and corrective
trajectories collected through human intervention and rollback
\cite{wu2026flowpro}. Across these approaches, corrective data collection
and repeated real-robot evaluation demand substantial expert time and
hardware resources, constraining systematic algorithm exploration.

\subsection{Task Monitoring and Failure Recovery}

Task monitoring connects execution feedback to failure detection and
recovery \cite{huang2023innermonologue,liu2023reflect}. DoReMi uses VLM
queries to detect task-constraint violations and trigger replanning
\cite{guo2024doremi}. Language-model agents also encode task knowledge in
executable programs for robot planning and control, as demonstrated by Code
as Policies \cite{liang2023codeaspolicies}, ProgPrompt
\cite{singh2023progprompt}, and VoxPoser \cite{huang2023voxposer}.
Code-as-Monitor extends this idea to execution monitoring by compiling task
constraints into code that checks tracked geometric elements, reducing
reliance on repeated VLM queries \cite{zhou2025codeasmonitor}. AgentChord
complements monitoring with context-aware recovery branches added to task
graphs before execution, allowing detected failures to trigger prepared
corrective behaviors \cite{xu2026agentchord}. We instead use an
Agent-Compiled Progress Graph to guide expert recovery from stable snapshots
after terminal rollout failure, generating corrective trajectories for VLA
post-training.

\subsection{Simulation for Data Generation and Policy Evaluation}

Simulation supports both scalable training-data generation
\cite{wang2024robogen,wang2024gensim} and controlled policy evaluation
\cite{james2020rlbench,gu2023maniskill2}. RoboTwin combines generative digital
twins with task code to generate demonstrations and benchmark manipulation
policies \cite{mu2025robotwin}. Related benchmarks study knowledge transfer
in lifelong robot learning, as in LIBERO \cite{liu2023libero}, and unify
policy evaluation across simulation and the real world, as in RoboDojo
\cite{chen2026robodojo}.

IntervenGen extends data generation to recovery behavior, focusing on
robustness to object pose estimation errors. It combines closed-loop policy
execution with transformed replay of human source interventions to
synthesize corrective trajectories \cite{hoque2024intervengen}. RLinf provides
scalable infrastructure for embodied RL \cite{yu2026rlinf}, while its
real-world extension, USER, supports asynchronous online policy learning.
Its HG-DAgger workflow relies on human experts to determine intervention
timing and provide corrective actions \cite{zang2026user}. Our simulation
framework supports intervention-based VLA post-training without human
takeover.

\section{Method}
\label{sec:method}

\subsection{Framework Overview}
\label{sec:framework}

As shown in Fig.~\ref{fig:framework}, SimPT uses a Progress Graph to connect
policy execution with automated corrective data collection, followed by
policy post-training and standalone evaluation. Language-model agents compile
task specifications into candidate graphs, which undergo validation and
selection. Scripts then calibrate the selected graph's data-dependent
thresholds using simulated expert trajectories. During policy execution,
the graph tracks task progress and retains stable snapshots. If the rollout
ends in failure, a programmatic expert attempts recovery from these snapshots
to generate corrective trajectories. The collected rollouts and corrections
support post-training under different learning objectives, while standalone
evaluation compares the resulting policies and guides algorithm development.
After validating selected refinements in simulation, we apply them to
real-robot HIL post-training, aiming to reduce the expert time and hardware
resources needed for algorithm development.

\subsection{Agent-Compiled Progress Graph}
\label{sec:progress_graph}

\begin{figure*}[!t]
    \centering
    \setlength{\abovecaptionskip}{1pt}
    \includegraphics[width=\textwidth]{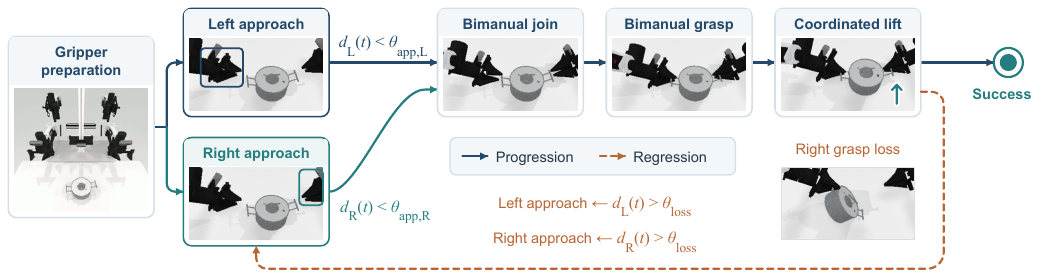}
    \caption{Selected components of the Lift Pot Progress Graph illustrate
    stage completion and regression after grasp loss. Solid arrows indicate
    progression, while dashed orange arrows indicate regression.
    The approach, join, grasp, and lift images show successive states from
    one rollout; the grasp-loss image comes from a separate failed rollout.
    Gripper preparation is shown from an external viewpoint to display both arms.}
    \label{fig:progress_graph}
\end{figure*}

The Agent-Compiled Progress Graph (PG) consists of a task-stage graph
$\mathcal{G}=(\mathcal{V},\mathcal{E})$ and transition conditions evaluated on
privileged simulator state $s_t$. Here, $\mathcal{V}$ is the set of task-stage
nodes, and $\mathcal{E}$ is the set of directed edges for progression and
regression. Forward edges activate subsequent stages when completion
conditions are satisfied, while backward edges reactivate earlier stages
when regression conditions are triggered.

To construct the graph, a preparatory agent reads the RoboTwin task source
code and produces a structured specification containing the task instruction,
object and arm assignments, available state fields, stage dependencies, and
success criterion. Task-specific simulator adapters expose state fields such
as object poses, gripper states, and contact measurements.

We pass this specification and a reusable library of state checks to
$K$ independent Codex agents based on GPT-5.6-Sol
(Fig.~\ref{fig:framework}A). Each agent compiles a candidate graph by assigning
checks to stages and defining transitions for stage completion and loss of
prerequisites. Candidate graphs undergo static validation against the
specification, and the agents automatically revise them in response to
validation feedback. Among valid candidates, we prefer fewer state fields,
then fewer stages, and finally fewer checks. The selected graph specifies
the task structure and symbolic checks, leaving data-dependent thresholds
for subsequent calibration.

Fig.~\ref{fig:progress_graph} illustrates the Progress Graph for Lift Pot.
Before bimanual grasping, each arm must approach its assigned contact point
on the pot. Separate branches track the left and right approach stages.
Let $d_L(t)$ and $d_R(t)$ denote the distances from the respective tool center
points to these contact points. The left and right approach stages are
complete when
\begin{equation}
    d_L(t)<\theta_{\mathrm{app},L},
    \qquad d_R(t)<\theta_{\mathrm{app},R},
    \label{eq:approach_completion}
\end{equation}
respectively, where $\theta_{\mathrm{app},a}$ is the approach-distance
threshold for arm $a\in\{L,R\}$. The branches merge at a join that verifies
completion of both approach stages and checks that both arms satisfy the
proximity conditions for bimanual grasping. Grasp loss during grasping or lifting can
trigger regression to earlier stages. During a coordinated lift, for example,
the right gripper may separate from the pot while the left gripper maintains
its grasp. The graph detects this loss using the geometric condition
$d_R(t)>\theta_{\mathrm{loss}}$, where $\theta_{\mathrm{loss}}$ is a fixed
prior specified in the graph. It then reactivates the right approach stage
and marks the dependent bimanual join, grasp, and lift stages as incomplete,
while preserving completion of the left approach stage.

Before data collection, scripts calibrate data-dependent thresholds,
such as $\theta_{\mathrm{app},L}$ and $\theta_{\mathrm{app},R}$, using simulated
expert trajectories. For the approach stages in Fig.~\ref{fig:progress_graph},
each threshold is estimated from an upper quantile of the distances measured
when the corresponding arm completes its approach. The grasp-loss threshold
$\theta_{\mathrm{loss}}$ remains fixed at its graph prior, while task-success
constants follow the original task definition. Using expert replays, the
agent diagnoses and revises stage-completion and regression conditions. The
graph is then recalibrated and validated through replay. The final graph and
its parameters remain fixed during data collection. At runtime, state checks
track task progress and guide recovery snapshot selection without further
agent queries.

\subsection{Snapshots and Expert Recovery}
\label{sec:recovery}

During policy execution, SimPT retains the latest snapshot recorded at each
task stage. Each snapshot stores the simulator state, active stages, and
capture step. We denote the set of retained snapshots at policy step $t$ by
\begin{equation}
    \mathcal{B}_t=\{\sigma_{\mathrm{init}},\sigma_1,\ldots,\sigma_{N_t}\},
    \label{eq:recovery_snapshots}
\end{equation}
where $\sigma_{\mathrm{init}}$ is the initial-state snapshot, $\sigma_i$
denotes a snapshot retained during policy execution, and $N_t$ is the number
of retained non-initial snapshots.

When a rollout ends in failure, SimPT attempts expert recovery from the
retained snapshots, starting with the most advanced task stages. For each
attempt, it restores the saved
simulator state and invokes a CuRobo-based programmatic expert to complete
the remaining task. If recovery succeeds, the resulting trajectory is
retained as corrective data. Otherwise, SimPT tries a snapshot from an
earlier task stage. The initial-state snapshot serves as the final fallback,
allowing the expert to restart the task.

\subsection{Post-Training and Policy Evaluation}
\label{sec:post_training}

SimPT records successful and failed policy rollouts, together with expert
corrective data, in a buffer for VLA post-training:
\begin{equation}
    \mathcal{D}
    = \mathcal{D}^{\pi}_{\mathrm{succ}}
    \cup \mathcal{D}^{\pi}_{\mathrm{fail}}
    \cup \mathcal{D}_{\mathrm{corr}},
    \label{eq:data_buffer}
\end{equation}
where $\mathcal{D}^{\pi}_{\mathrm{succ}}$ and
$\mathcal{D}^{\pi}_{\mathrm{fail}}$ contain successful and failed rollouts
of the VLA policy $\pi$, respectively, and $\mathcal{D}_{\mathrm{corr}}$
contains expert corrective data. Each method selects data from this buffer
and constructs training samples according to its learning objective.

For methods that require complete trajectories, SimPT constructs a corrected
trajectory $\tau^{+}$ from a failed rollout $\tau^{-}$ by appending the expert
recovery segment to the policy prefix before recovery step $t$:
\begin{equation}
    \begin{aligned}
        \tau^{-} &= \tau^{-}_{<t} \oplus \tau^{-}_{\ge t},\\
        \tau^{+} &= \tau^{-}_{<t} \oplus \tau^{\mathrm{E}}_{t}.
    \end{aligned}
    \label{eq:corrected_trajectory}
\end{equation}
Here, $\oplus$ denotes concatenation, and $\tau^{-}_{<t}$ is the policy prefix
shared by the failed rollout and the complete corrected trajectory. The
suffixes $\tau^{-}_{\ge t}$ and $\tau^{\mathrm{E}}_{t}$ represent the failed
policy continuation and successful expert recovery from the restored state,
respectively. After post-training, standalone evaluation measures the
updated policy's performance.

\section{Experiments}
\label{sec:experiments}

\subsection{Experimental Setup}
\label{sec:experimental_setup}

To evaluate SimPT, we first compare PG with the fixed-interval Naive baseline
for corrective data collection. We then implement and evaluate RECAP
\cite{physicalintelligence2025recap}, HG-DAgger, DPO+SFT, and FlowPRO on seven
RoboTwin manipulation tasks (Table~\ref{tab:recovery_results})
\cite{mu2025robotwin}: Lift Pot, Move Can Pot, Place Empty Cup, Beat Block
Hammer, Place Can Basket, Handover Block, and Stack Blocks Two. These tasks
cover bimanual coordination, precise placement, tool use, and multistage
manipulation. DPO+SFT and FlowPRO form preference pairs from failed and
corrective trajectories. Following Appendix~J of FlowPRO
\cite{wu2026flowpro}, DPO+SFT combines a flow-matching DPO loss with a
supervised fine-tuning (SFT) objective, while FlowPRO adds a proximal
regularizer. Finally, we use SimPT to refine RECAP in simulation and validate
the refinement on a real-robot cable-insertion task.

For each task, Codex (GPT-5.6-Sol) generates three candidate graphs and
performs up to two revisions per candidate using static-validation feedback.
All policies use the $\pi_{0.5}$ backbone and an action chunk size of 50.
We train a parent policy for each task using SFT on 50 expert demonstrations.
Naive (N) and Progress Graph (PG) collection use the same parent policy
checkpoint, 240 prescribed environment seeds, and CuRobo-based expert, with
rollouts executed independently for each procedure. The Naive baseline saves
snapshots every 50 policy steps and attempts recovery in reverse chronological
order after task failure.

For each task and collection procedure, the methods share a fixed dataset
collected with the SFT parent
policy and construct training samples according to their learning objectives.
All methods use 10,000 policy training steps. For each task, we use 40
validation seeds to tune parameters such as the RECAP guidance strength.
Final held-out evaluation uses 200 fixed environment seeds per task that pass
the official expert's planning and task-success checks. These seeds are
disjoint from the collection and validation sets, giving 1,400 trials per
method across seven tasks. Standalone evaluation runs only the updated VLA,
without the PG, recovery expert, or privileged simulator state.

\begin{figure*}[t]
    \centering
    \setlength{\abovecaptionskip}{1pt}
    \includegraphics[width=\textwidth]{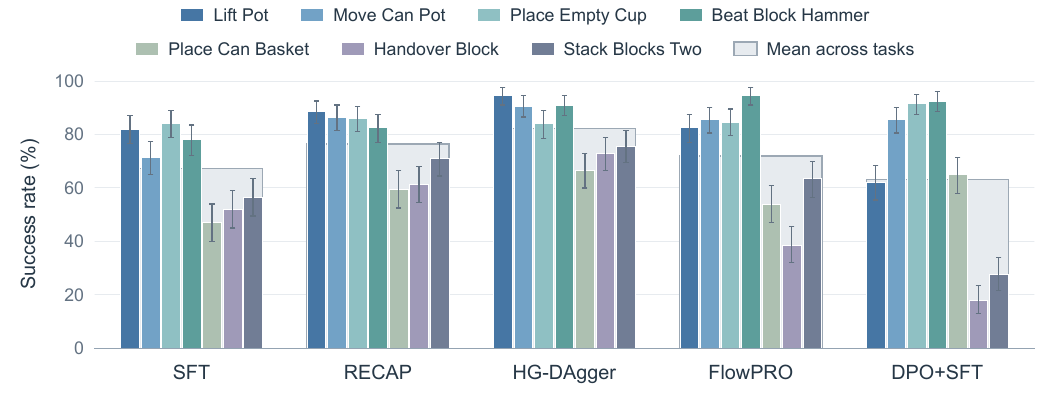}
    \caption{Four post-training methods are compared with the SFT baseline
    on seven RoboTwin tasks. All post-training methods use data from
    Progress Graph (PG) collection. Standalone policy evaluation uses 200
    held-out seeds per task, totaling 1,400 trials per method. Methods form
    groups on the horizontal axis, with seven colored bars per group
    corresponding to the tasks in legend order. Bar heights show task success
    rates; wide light-gray background bars span each group and show
    equal-weight means across tasks. Vertical whiskers indicate 95\%
    bootstrap confidence intervals over evaluation seeds (20,000 resamples),
    conditional on the trained policies.}
    \label{fig:policy_results}
\end{figure*}

\subsection{Progress Graph Construction and Recovery}
\label{sec:recovery_results}

The agent produces candidate graphs that pass static validation for all
seven tasks, with a final pass rate of 100\% (21/21). Nine candidates
(42.9\%) pass on the first attempt. The remaining 12 require an average of
1.17 revisions, which the agent performs automatically using validation
feedback. Graph generation and static validation take an average
of 8.0 minutes per task, ranging from 4.5 to 11.5 minutes.

We compare PG and Naive in terms of search efficiency and recovery success.
Each snapshot tested by the expert counts as one recovery attempt. We measure
search efficiency by the total number of attempts and the mean attempts per
successful recovery, computed as total attempts divided by recovered
rollouts. Recovery success rate is the proportion of failed policy rollouts
that the expert successfully recovers.

\begin{table*}[t]
    \centering
    \caption{\parbox[t]{\textwidth}{\upshape Recovery success and search
    efficiency of independent fixed-interval Naive (N) and Progress Graph
    (PG) collections across seven RoboTwin tasks. Attempts per recovery equal
    total attempts divided by recovered rollouts. Reduction columns report the
    relative decrease from N to PG using unrounded values; aggregate metrics
    use pooled counts. Bold indicates the better or tied N/PG result.}}
    \label{tab:recovery_results}
    \footnotesize
    \setlength{\tabcolsep}{4pt}
    \begin{tabular*}{\textwidth}{@{\extracolsep{\fill}}ccccccccc@{}}
        \toprule
        \multirow{2}{*}{Task}
        & \multicolumn{3}{c}{Total attempts $\downarrow$}
        & \multicolumn{3}{c}{Attempts per recovery $\downarrow$}
        & \multicolumn{2}{c}{Recovery success (\%) $\uparrow$} \\
        \cmidrule(lr){2-4}\cmidrule(lr){5-7}\cmidrule(lr){8-9}
        & N & PG & Reduction (\%)
        & N & PG & Reduction (\%)
        & N & PG \\
        \midrule
        Lift Pot
        & 238 & \textbf{63} & 73.5
        & 4.58 & \textbf{1.21} & 73.5
        & \textbf{96.3} & \textbf{96.3} \\
        \addlinespace[3pt]
        Move Can Pot
        & 180 & \textbf{87} & 51.7
        & 3.46 & \textbf{1.55} & 55.1
        & 94.5 & \textbf{96.6} \\
        \addlinespace[3pt]
        Place Empty Cup
        & 89 & \textbf{46} & 48.3
        & 2.70 & \textbf{1.12} & 58.4
        & \textbf{100.0} & \textbf{100.0} \\
        \addlinespace[3pt]
        Beat Block Hammer
        & 76 & \textbf{66} & 13.2
        & 1.58 & \textbf{1.40} & 11.3
        & \textbf{100.0} & \textbf{100.0} \\
        \addlinespace[3pt]
        Place Can Basket
        & 1,295 & \textbf{345} & 73.4
        & 10.88 & \textbf{2.72} & 75.0
        & 93.7 & \textbf{100.0} \\
        \addlinespace[3pt]
        Handover Block
        & 1,329 & \textbf{192} & 85.6
        & 11.87 & \textbf{1.63} & 86.3
        & 83.6 & \textbf{98.3} \\
        \addlinespace[3pt]
        Stack Blocks Two
        & 574 & \textbf{146} & 74.6
        & 7.17 & \textbf{1.68} & 76.6
        & \textbf{100.0} & \textbf{100.0} \\
        \midrule
        \textbf{Aggregate}
        & 3,781 & \textbf{945} & 75.0
        & 7.62 & \textbf{1.79} & 76.5
        & 93.4 & \textbf{98.9} \\
        \bottomrule
    \end{tabular*}
\end{table*}

Table~\ref{tab:recovery_results} shows that PG requires fewer recovery
attempts than Naive on all seven tasks while maintaining or improving
recovery success rates. Across tasks, PG reduces the total attempt count
from 3,781 to 945, a reduction of 75.0\%. The mean number of attempts per
successful recovery falls from 7.62 to 1.79, while the overall recovery
success rate rises from 93.4\% to 98.9\%.
Handover Block shows the largest reduction in recovery attempts, from
1,329 to 192, together with an increase in recovery success rate from
83.6\% to 98.3\%. These results show more efficient recovery search under PG
collection. We next compare the policies trained on the resulting datasets.

\subsection{Comparison of Post-Training Methods}
\label{sec:policy_results}

We first compare the four post-training methods using PG-collected data,
with SFT as the baseline (Fig.~\ref{fig:policy_results}). HG-DAgger achieves
the highest overall success rate of 82.14\% and performs best on five of the
seven tasks. Its behavior-cloning objective directly supervises the policy
with expert corrective actions. RECAP achieves 76.50\%, exceeding the SFT
baseline of 67.29\%. Its advantage-conditioned training relies on value
estimation \cite{physicalintelligence2025recap}. With limited trajectory data,
the value function can overfit, and temporal fluctuations in its predictions
can compromise the reliability of advantage labels.

FlowPRO and DPO+SFT achieve overall success rates of 71.86\% and 63.14\%,
respectively, both below RECAP; DPO+SFT also falls below SFT.
Nevertheless, they perform best on Beat Block Hammer and Place Empty Cup,
respectively.
Relative to DPO+SFT, FlowPRO adds a proximal regularizer that constrains
the implicit reward magnitudes of preferred and rejected actions to
mitigate reward hacking and stabilize preference optimization
\cite{wu2026flowpro}. Both methods synthesize preferred action chunks by
interpolating toward successful recovery trajectories. This interpolation
does not account for object configurations or contact states, so action
feasibility at failure states is not guaranteed.

\begin{table}[!htb]
    \centering
    \caption{\parbox[t]{\columnwidth}{\upshape Overall policy success rates are compared after post-training
    with data from Naive or Progress Graph (PG) collection. Each method is
    evaluated on 1,400 held-out trials per collection procedure across
    seven RoboTwin tasks. $\Delta$ denotes PG minus Naive in percentage
    points. Bold indicates the higher success rate within each row.}}
    \label{tab:policy_results}
    \footnotesize
    \setlength{\tabcolsep}{4pt}
    \begin{tabular*}{\columnwidth}{@{\extracolsep{\fill}}lrrr@{}}
        \toprule
        Method & Naive (\%) & PG (\%) & $\Delta$ (pp) \\
        \midrule
        RECAP & 66.64 & \textbf{76.50} & $+9.86$ \\
        HG-DAgger & 80.21 & \textbf{82.14} & $+1.93$ \\
        DPO+SFT & 62.64 & \textbf{63.14} & $+0.50$ \\
        FlowPRO & \textbf{76.07} & 71.86 & $-4.21$ \\
        \bottomrule
    \end{tabular*}
\end{table}

We next compare each method using data from Naive and PG collection
(Table~\ref{tab:policy_results}). PG collection improves the overall success
rates of RECAP, HG-DAgger, and DPO+SFT, with the largest gain for RECAP.
RECAP achieves 66.64\% under Naive collection, slightly below SFT, and
76.50\% under PG collection, a gain of
9.86 percentage points (pp), with improvements on all seven tasks. HG-DAgger
and DPO+SFT gain 1.93 and 0.50~pp, respectively.
FlowPRO is the exception: its success rate decreases from 76.07\% under
Naive collection to 71.86\% under PG collection, a drop of 4.21~pp.
Overall, PG-collected data improve policy success rates for most of the
evaluated post-training methods.

\subsection{Algorithm Exploration and Real-Robot Validation}
\label{sec:recipe_case}

Using SimPT, we develop \emph{RECAP-CF-LPF} to refine the value-based
advantage labels used by RECAP. Standard RECAP fits a value model to the
collected trajectories and constructs advantage labels from predictions on
those same trajectories, making the labels susceptible to value-model
overfitting. We mitigate this issue with five-fold cross-fitting (CF),
predicting values for each trajectory with a model trained on the other
folds~\cite{chernozhukov2018dml}. To attenuate high-frequency noise before
computing 50-step advantages, we apply a second-order, zero-phase Butterworth
low-pass filter (LPF)~\cite{butterworth1930filter} with a 1~Hz cutoff to the
out-of-fold value sequence sampled at 30~Hz. We compare this combined
refinement with standard RECAP using PG-collected data from the seven tasks,
then apply it to real-robot HIL post-training.

\begin{figure}[!t]
    \centering
    \setlength{\abovecaptionskip}{3pt}
    \includegraphics[width=\columnwidth]{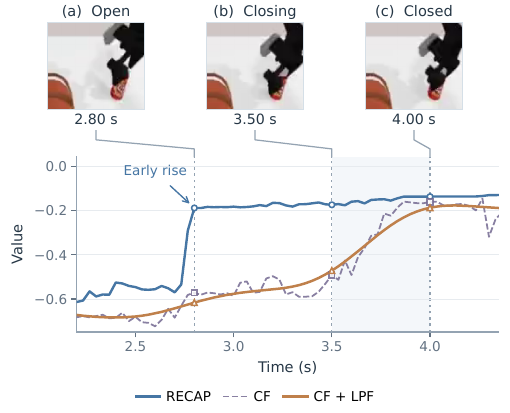}
    \caption{Value sequences during grasping in a Place Can Basket expert recovery
    trajectory. Curves show standard RECAP, out-of-fold predictions from
    cross-fitting (CF), and the CF sequence after 1~Hz Butterworth low-pass
    filtering (RECAP-CF-LPF). Images (a--c) show the gripper open,
    closing, and closed, with dashed lines linking each image to its
    timestamp. The shaded interval spans the closing and closed gripper
    states (3.50--4.00~s).}
    \label{fig:recap_value_grasp_timing}
\end{figure}

\begin{figure}[!t]
    \centering
    \setlength{\abovecaptionskip}{3pt}
    \includegraphics[width=\columnwidth]{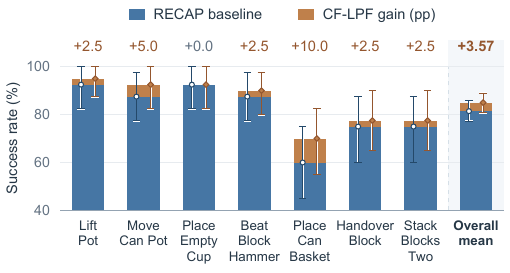}
    \caption{Validation success rates on seven RoboTwin tasks using PG
    collection and 40 seeds per task. Blue bars show standard RECAP, and
    orange segments show the gain from RECAP-CF-LPF (five-fold cross-fitting
    and 1~Hz Butterworth low-pass filtering). Bar tops give RECAP-CF-LPF
    success rates; labels report gains in percentage points. Circles and
    diamonds mark standard RECAP and RECAP-CF-LPF estimates, respectively.
    Whiskers show 95\% bootstrap confidence intervals over validation seeds
    for the selected policies (20,000 resamples). The vertical axis starts
    at 40\%.}
    \label{fig:recap_validation}
\end{figure}

Fig.~\ref{fig:recap_value_grasp_timing} compares value sequences during
grasping in one Place Can Basket recovery trajectory. The standard RECAP
prediction rises before gripper closure, whereas the CF
prediction rises mainly during closure. In this example, the CF prediction
is intuitively more plausible because its rise aligns more closely with the
grasping motion. Applying LPF to the CF sequence attenuates local fluctuations
before advantage computation.

Fig.~\ref{fig:recap_validation} shows that RECAP-CF-LPF increases the overall
validation success rate from 81.43\% to 85.00\%, a gain of 3.57~pp.
Success rates improve on six tasks and remain unchanged on
Place Empty Cup. Place Can Basket shows the largest gain, from 60.0\% to 70.0\%.

On the held-out evaluation set, RECAP-CF-LPF achieves 79.36\% overall
success, compared with 76.50\% for standard RECAP, a gain of 2.86~pp across
1,400 trials per variant. Success rates increase on five tasks and decrease
by 1.0~pp on Beat Block Hammer and 2.0~pp on Handover Block. The overall
improvement has a 95\% confidence interval of 0.36--5.36~pp, estimated using
20,000 paired bootstrap resamples of evaluation seeds within each
task~\cite{efron1993bootstrap}.

\begingroup
\widowpenalty=10000 
To test whether the refinement developed in simulation improves real-robot
HIL post-training, we evaluate RECAP-CF-LPF on an Ethernet cable-insertion
task. The platform comprises two AgileX PiPER arms with grippers and three
RGB cameras: one head camera and one wrist camera per arm. Human experts
operate two corresponding master arms to collect demonstrations and
corrections. Both variants use 1,672 training
trajectories: 200 human demonstrations, 802 episodes with human
interventions, and 670 fully autonomous policy rollouts.
\par
\endgroup

\begin{figure}[!t]
    \centering
    \setlength{\abovecaptionskip}{3pt}
    \includegraphics[width=\columnwidth]{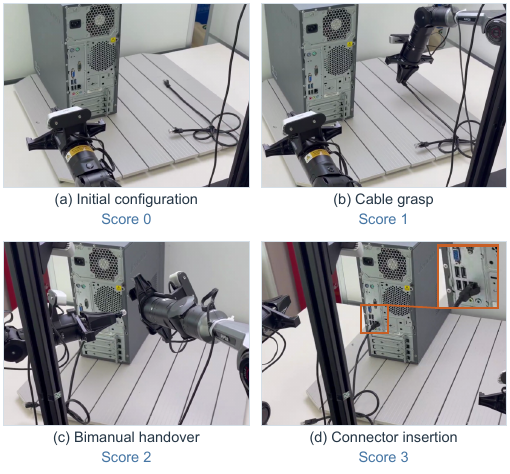}
    \caption{Real-robot Ethernet cable insertion is evaluated using a
    cumulative progress score. Panels (a--d) show the initial configuration,
    cable grasping, bimanual handover, and plug insertion, corresponding to
    scores of 0, 1, 2, and 3. Each completed milestone adds one point, and
    full insertion defines task success. The inset in (d) provides a
    close-up of the inserted plug.}
    \label{fig:real_robot_task}
\end{figure}

As illustrated in Fig.~\ref{fig:real_robot_task}, the robot grasps the cable,
transfers it between arms, and aligns and inserts the plug into a desktop
computer's Ethernet port. The progress score starts at zero and increases by
one after each milestone: cable grasping, bimanual handover, and plug
insertion. Full insertion defines task success and yields the maximum score
of three.

\begingroup
\widowpenalty=10000
We evaluate 25 distinct scenarios varying cable length, initial position,
and computer chassis placement. RECAP-CF-LPF succeeds in 24/25 scenarios
(96\%), compared with 18/25 (72\%) for standard RECAP. All 18 scenarios
solved by standard RECAP are also solved by RECAP-CF-LPF. The corresponding
cumulative progress scores are 72/75 and 59/75, respectively,
out of a maximum of 75 points across the 25 scenarios.
These results show that the refinement developed with SimPT improves policy
performance when applied to real-robot HIL post-training.
\par
\endgroup

\section{Conclusion and Future Work}
\label{sec:conclusion}

We presented SimPT, a simulation framework for systematically exploring
intervention-based VLA post-training algorithms. It automates policy rollouts,
corrective data collection, policy training, and standalone evaluation. Its
Agent-Compiled Progress Graph guides expert recovery after rollout failure,
enabling corrective data collection without human takeover. Across seven
RoboTwin tasks, PG achieved higher recovery success with fewer attempts than
a fixed-interval baseline, and PG-collected data improved policy success for
most evaluated methods. Using SimPT, we developed RECAP-CF-LPF, which
combines cross-fitting and low-pass filtering to refine RECAP's value-based
advantage labels. The resulting gains in simulation and real-robot HIL
post-training demonstrate SimPT's utility for algorithm exploration and
development.

Post-training performance remains sensitive to task and scene
characteristics. We plan to incorporate real-to-sim reconstruction
\cite{torne2024rialto,han2026re3sim}
to better align simulated tasks and scenes with deployment conditions,
strengthening SimPT as a platform for developing and evaluating real-robot
HIL post-training methods.

\bibliographystyle{IEEEtran}
\bibliography{references}

\end{document}